\documentclass[]{spie}

\usepackage{amsmath,amssymb}
\usepackage{graphicx}
\usepackage{booktabs}
\usepackage{multirow}
\usepackage{xcolor}
\usepackage{hyperref}

\newcommand{\best}[1]{\textbf{#1}}

\title{Taxonomy-aware deep learning for hierarchical marine species classification in underwater imagery}

\author[a]{Dan Zimmerman}
\author[a]{Dimitris A. Pados}
\author[a]{George Sklivanitis}

\affil[a]{Center for Connected Autonomy and AI}
\affil[]{Florida Atlantic University, Boca Raton, FL 33431, USA}

\authorinfo{Further author information: (Send correspondence to D.Z.)\\
D.Z.: E-mail: dzimmerman2021@fau.edu\\
D.P.: E-mail: dpados@fau.edu\\
G.S.: E-mail: gsklivanitis@fau.edu}

\begin{document}

\maketitle

\begin{abstract}
Automated classification of marine species from underwater imagery is essential for scalable ocean biodiversity monitoring and conservation policy.
Existing approaches struggle with severe domain shift across collection platforms, fine-grained visual similarity between closely related species, and uneven annotation granularity, where many specimens can only be identified to genus or a coarser taxonomic rank.
We present a taxonomy-aware deep learning framework that aligns both the training loss and the inference rule with the hierarchical structure of biological classification, combining a taxonomy-weighted loss, minimum-risk Bayesian inference, multi-scale feature encoding, and independent per-rank classification heads.
Evaluated on the FathomNet 2025 dataset\cite{fathomnet2025kaggle} (79 marine classes across seven taxonomic ranks), the system achieves a mean taxonomic distance of 1.581, within 3\% of the 1st-place solution (1.535), 
with the largest gains from metric-aligned inference and simple, decoupled components that generalize better than learned dependencies under distribution shift.
\end{abstract}


\section{INTRODUCTION}

Marine biodiversity monitoring is fundamental to understanding ocean ecosystems, tracking the effects of climate change, and guiding conservation policy.\cite{katija2022fathomnet}
Underwater camera systems deployed on remotely operated vehicles, autonomous underwater vehicles, and fixed observatories now generate millions of images annually, far exceeding the capacity of human experts to annotate.
Automated visual classification of marine organisms is therefore essential, but presents challenges distinct from terrestrial recognition: severe domain shift across collection platforms, fine-grained visual similarity between closely related species, and uneven annotation granularity, where many specimens can only be identified to genus or a coarser rank rather than to species.
In biology, organisms are organized into a taxonomy, a hierarchical classification system with ranks from broad (kingdom, phylum) to specific (genus, species).

When species-level identification is uncertain, predicting at a coarser rank (genus, family) is vastly preferable to an incorrect species prediction.
This motivates taxonomy-aware classification, where prediction quality is measured not by flat accuracy but by the severity of the error within the biological hierarchy.
Standard cross-entropy training is misaligned with this objective: it treats all misclassifications equally, whether the confusion is between species in the same genus or between organisms in different phyla.

We present a taxonomy-aware deep learning framework that directly addresses this misalignment by aligning both the training loss and the inference rule with the hierarchical evaluation metric.
The framework combines a taxonomy-weighted loss, minimum-risk Bayesian inference, multi-scale feature encoding, and independent per-rank classification heads.
Two design principles produce disproportionate gains: (1)~metric-aligned training and inference account for roughly 11\% of the total improvement, with minimum-risk inference providing the most consistent gain; and (2)~simple, decoupled components consistently outperform learned dependencies under distribution shift.
Built on DINOv2-Base\cite{oquab2024dinov2} with layer-wise learning rate decay,\cite{howard2018ulmfit} the system achieves a mean taxonomic distance of 1.581 on the FathomNet 2025 dataset,\cite{fathomnet2025kaggle} within 3\% of the 1st-place solution (1.535).
A backbone comparison against ConvNeXtV2-Base\cite{woo2023convnextv2} (1.782) further isolates the importance of fine-tuning recipe: naive uniform fine-tuning of DINOv2 yields 2.225, and only layer-wise decay recovers the gains of self-supervised pretraining.

The same problem class, fine-grained classification under sensor and environmental domain shift with hierarchically structured label costs, recurs across automatic target recognition (ATR) and intelligence, surveillance, and reconnaissance (ISR) pipelines, where decision-theoretic inference over taxonomies of vehicles, vessels, or threats faces analogous challenges. The empirical findings reported here therefore inform a broader class of imaging applications constrained by similar data regimes.

\section{RELATED WORK}

Several families of methods have been proposed to exploit hierarchical label structure in classification.
\emph{Structural constraint} approaches enforce consistency between predictions at different levels of the hierarchy: Hierarchical Cross-Entropy (HXE)\cite{bertinetto2020making} decomposes the softmax likelihood along the class tree via conditional probabilities at each rank, and C-HMCNN\cite{giunchiglia2020coherent} uses a max-constraint module to guarantee that parent-class probabilities exceed those of their children.
\emph{Branching} architectures such as B-CNN\cite{zhu2017bcnn} attach coarse-level classifiers to earlier layers of the backbone, matching feature abstraction to the granularity of each rank.
A third strategy parameterizes only the finest-level (leaf) distribution and recovers coarse-level probabilities by marginalization over descendants;\cite{su2021bilinear} this avoids explicit inter-level coupling but forgoes direct supervision at coarser ranks.

Our approach falls into a different tradition: \emph{cost-sensitive classification},\cite{duda2001pattern} where the loss function encodes the relative severity of different errors.
Rather than enforcing structural constraints or factoring the predictive distribution, our taxonomic distance loss directly penalizes the expected value of the evaluation metric, and minimum-risk Bayesian inference selects predictions that minimize expected cost at test time.
We additionally find that simple feature concatenation across spatial scales generalizes better than learned attention-based fusion under distribution shift (four variants --- hierarchy-aware, patch self-attention, part-based, and combined --- each 7.3--16.9\% worse on test despite better cross-validation scores), consistent with recent observations that simpler fusion is more robust to domain mismatch.\cite{nguyen2024saft}
Concurrent work on the FathomNet 2025 dataset includes MATANet,\cite{lee2026matanet} which combines multi-context attention with hierarchical separation learning on DINOv2 backbones to achieve first place, and a fourth-place solution\cite{health9819} that also adopts minimum-risk inference.

\section{PROBLEM FORMULATION}

We consider the problem of hierarchical classification of marine organisms from underwater imagery. Marine organisms may belong to any of the following $7$ biological taxonomic ranks (Kingdom $\to$ Phylum $\to$ Class $\to$ Order $\to$ Family $\to$ Genus $\to$ Species).
Given an image region containing a marine organism, the challenge is to develop a model that can accurately classify its leaf-level taxon and ideally leverage its taxonomic information to do so.
Let $s$ denote the true leaf taxon and $\hat{s}$ the predicted leaf taxon.
Model performance is evaluated by the taxonomic distance between $\hat{s}$ and $s$, defined as the total number of tree edges traversed from $\hat{s}$ to $s$ through their lowest common ancestor\cite{aho1976lca} (LCA):
\begin{equation}
d(\hat{s}, s) = \text{depth}(\hat{s}) + \text{depth}(s) - 2\,\text{depth}(\text{LCA}(\hat{s}, s)),
\label{eq:distance}
\end{equation}
where $\text{depth}(\cdot)$ returns the level of a node in the taxonomic tree (i.e., $0$ for root, $1$ for kingdom, $7$ for species), and $\text{LCA}(\hat{s}, s)$ is the deepest node that is an ancestor of both $\hat{s}$ and $s$.
This distance is zero for an exact match and accommodates predictions at any taxonomic rank: a coarser prediction trades exact-match precision for a bounded penalty.
We average taxonomic distance across all $N$ test organisms, to calculate a $\text{Score} = \tfrac{1}{N}\sum_{n=1}^N d(\hat{s}_n, s_n)$.

This metric creates a fundamental misalignment with cross-entropy. For example, confusing species within the same genus ($d = 2$) is six times less costly than confusing across phyla ($d = 12$), yet cross-entropy treats both equally.
A defining additional challenge is distribution shift between training and deployment. Underwater imagery collected across different geographic regions, depths, and camera systems exhibits substantial variation in organisms, scale, lighting, and annotation density, motivating techniques that generalize robustly over those that may overfit to the training distribution.

\section{METHOD}

\begin{figure}[t]
\centering
\includegraphics[width=\textwidth]{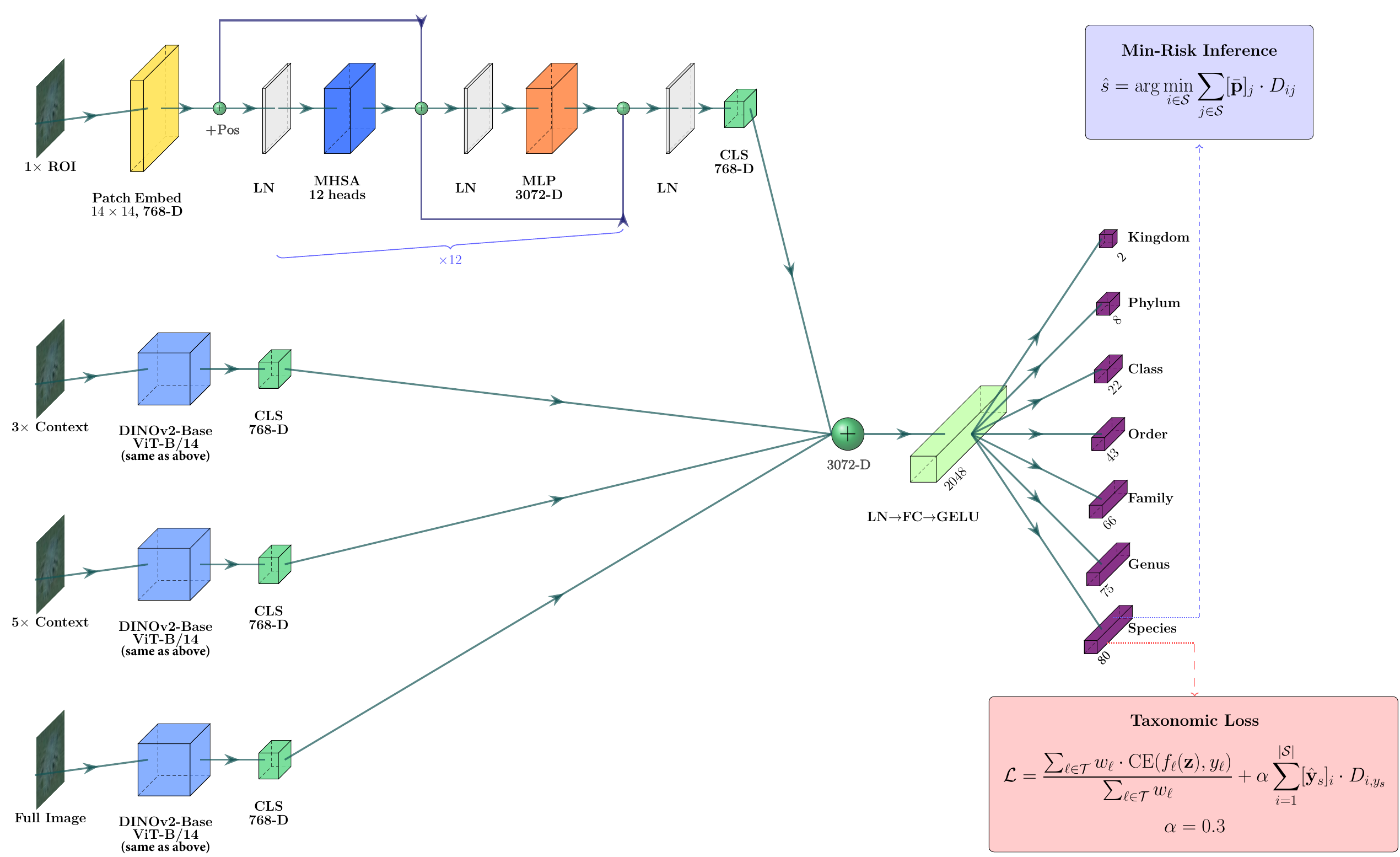}
\vspace{3pt}
\caption{DINOv2-Base ViT-B/14 architecture. Four crops at 1$\times$ (ROI), 3$\times$, 5$\times$, and full-image scales are processed by independent ViT-B/14 backbones (86M parameters each, $\sim$344M across all four scales); the 768-D CLS tokens are concatenated (3072-D), projected to 2048-D through a fusion layer (6.3M), and fed to seven independent classification heads (0.2M total). Full model: $\sim$351M parameters. The taxonomy-aware loss combines cross-entropy with expected taxonomic distance ($\alpha = 0.3$); minimum-risk inference selects the species minimizing expected distance.}
\label{fig:arch}
\end{figure}

We propose a taxonomy-aware deep-learning framework for hierarchical marine species classification comprising: 1) multi-scale feature extraction; 2) per-level classification heads at all seven taxonomic ranks; 3) a taxonomy-aware loss function; and 4) minimum-risk Bayesian inference over a cross-validation ensemble.
Figure~\ref{fig:arch} depicts the end-to-end pipeline.

\subsection{Multi-Scale Context Encoding}
\begin{figure}[t]
\centering
\includegraphics[width=\textwidth]{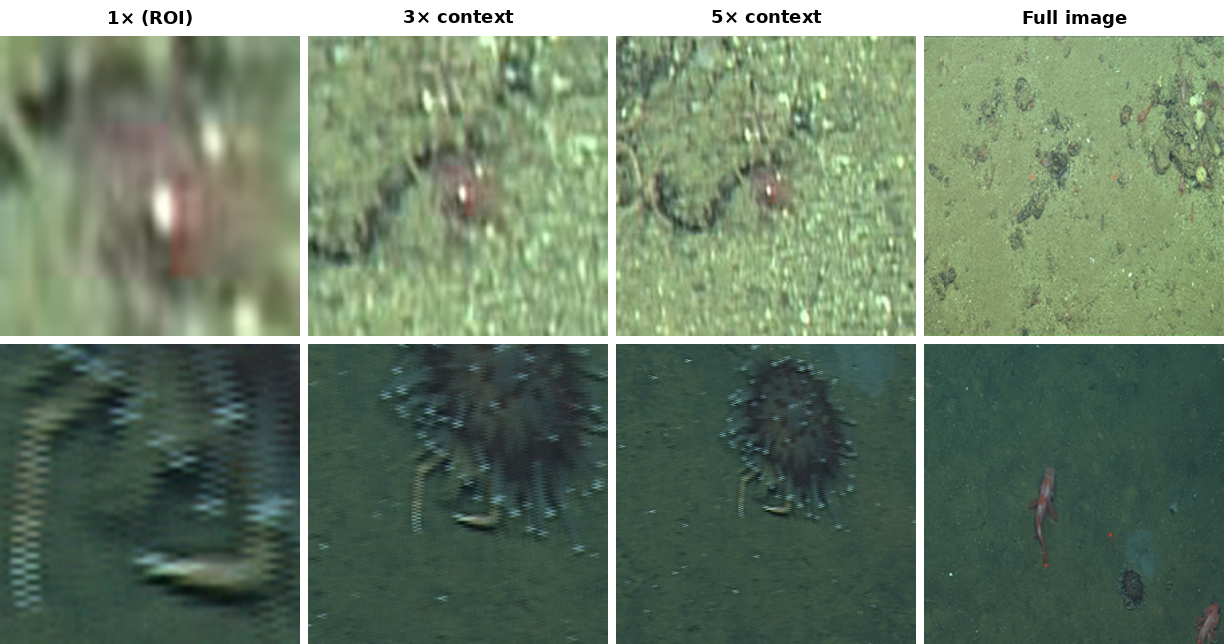}
\caption{Multi-scale context crops for two test specimens. Each row shows the same organism at 1$\times$ (tight ROI), 3$\times$, 5$\times$, and full-image scales. Increasing context reveals substrate type, co-occurring organisms, and environmental cues that aid classification, while the ROI crop preserves fine-grained morphological detail.}
\label{fig:multiscale}
\end{figure}

 As illustrated in Figure~\ref{fig:arch}, the model takes as input not only the region of interest (ROI) containing the target marine organism, but also contextual regions at 3$\times$, 5$\times$, and full-image scales, all center-cropped for the ROI. Figure~\ref{fig:multiscale} depicts each one of the four scaled images for two test marine organisms.
When expanded crops exceed image boundaries, coordinates are clamped to preserve real content rather than padding artifacts.
All crops are resized to $224 \times 224$ and normalized using ImageNet statistics.
Each scale is processed by an independent DINOv2-Base\cite{oquab2024dinov2} backbone (ViT-B/14, 768-D CLS-token output). Specifically, each ViT encoder extracts the classification (CLS) token embedding $g$ from the marine organism ROI and the patch embeddings $p_{cr}$ from each context region $c_r$ to form an output feature vector $\mathbf{z}_c \in \mathbb{R}^{768}$.  
The four output feature vectors at each context level are then concatenated with the ROI embedding along the embedding dimension to form an embedding vector $\mathbf{z}_{mc} \in \mathbb{R}^{3072}$, and passed through a normalization layer and a projection layer $\text{proj}(\cdot)$
 to produce a final fused embedding vector $\mathbf{z} \in \mathbb{R}^{2048}$. A GELU (Gaussian Error Linear Unit) activation function and a dropout layer are subsequently applied to the projected embedding.

\subsection{Hierarchical Classification Heads}

Marine animals show different visual characteristics across the seven taxonomic levels. We maintain seven classification heads, one per taxonomic level, each operating independently on the fused embedding vector $\mathbf{z}$. Higher-level categories rely on global shape and morphology, whereas lower-level categories depend more on subtle, fine-grained features. 
Coupling heads via parent conditioning (each head receiving its parent's $\text{softmax}(\cdot)$ as additional input) slightly degrades performance under distribution shift (as we will discuss in Section~\ref{sec:ablation}), so we adopt independent classification as the default.
For each level $\ell$, an individual auxiliary classifier $f_{\ell}(\cdot)$ is attached, and the ground-truth label $y_{\ell}$ at that level is used to compute the cross-entropy loss.

\subsection{Taxonomy-Aware Objective Function}
\label{sec:taxloss}

We propose to combine a weighted cross-entropy loss across all levels with a taxonomic distance penalty at the species level. The complete objective function is therefore written as
\begin{equation}
\mathcal{L} = \frac{\sum_{\ell \in \mathcal{T}} w_\ell \cdot \text{CE}(f_{\ell}({\bf z}), y_\ell)}{\sum_{\ell \in \mathcal{T}} w_\ell} + \alpha \cdot \sum_{i=1}^{|\mathcal{S}|} \hat{y}_{\rm{species}}(i) \cdot D_{i, y_{\ell|\ell={\rm{species}}}},
\label{eq:loss}
\end{equation}
where $\mathcal{T} = \{\text{kingdom}, \text{phylum}, \ldots, \text{species}\}$ is the set of seven taxonomic levels, 
$w_\ell$ is the weight for each level $\ell$, and $\text{CE}$ denotes cross-entropy with label smoothing ($\epsilon = 0.1$).
In the second term, $\mathcal{S}$ is the set of leaf-level taxa (species head), $\hat{\mathbf{y}}_{\rm{species}} \in \mathbb{R}^{|\mathcal{S}|}$ is the species-head softmax output vector, $\hat{y}_{\rm{species}}(i)$ denotes its $i$-th component (the predicted probability of leaf taxon $i$), $D \in \mathbb{R}^{|\mathcal{S}| \times |\mathcal{S}|}$ is the precomputed pairwise tree-path distance matrix (normalized to $[0, 1]$; see below), and $\alpha = 0.3$ controls the trade-off between the cross-entropy and distance terms.
The weights-per-level $w_\ell$ increase from coarse to fine ranks (kingdom: 0.5, phylum: 0.75, class: 1.0, order: 1.25, family: 1.5, genus: 2.0, species: 2.5). Since the first term in Equation~\ref{eq:loss} is normalized by $\sum_{\ell \in \mathcal{T}} w_\ell$, only the ratios between weights affect training.

We set $|\mathcal{S}| = 80$ leaf labels (33 species, 46 coarse taxa identified from genus to phylum, and 1 ``unknown'' class) as a label set, and define $D \in \mathbb{R}^{80 \times 80}$ as the pairwise matrix of taxonomic distances over all leaves, with entries $D_{ij}$ computed by Eq.~\ref{eq:distance}. The training loss therefore descends directly on the evaluation metric, applied to the flat leaf label set $\mathcal{S}$.

The matrix is normalized to $[0, 1]$ by dividing by the observed maximum pairwise distance ($D_{\max} = 14$, corresponding to two leaves in different kingdoms).
This makes $y_s$ well-defined for every sample (i.e., it is always the assigned leaf) and the expected-distance loss applies uniformly, without per-sample masking or descendant aggregation.
The ``unknown'' class is placed under a designated parent at each level; its distances to all named taxa are set to $D_{\max}$, so confusing ``unknown'' with a named taxon (or vice versa) incurs the maximum penalty.
The taxonomic distance term can be viewed as a form of {structured label smoothing}\cite{muller2019labelsmoothing}: whereas standard label smoothing distributes $\epsilon$ uniformly, our term distributes it proportionally to taxonomic proximity.

\subsection{Ensemble and Minimum-Risk Inference}
\label{sec:ensemble}

We train a $K$-fold stratified cross-validation ensemble using identical hyperparameters across folds and average species-level $\text{softmax}$ probabilities during inference: ${\mathbf{p}} = \tfrac{1}{K}\sum_{k=1}^{K} \hat{\mathbf{y}}_{\rm{species}}^{(k)}$, where $K = 10$ is the number of ensemble members and $\hat{\mathbf{y}}_{\rm{species}}^{(k)} \in \mathbb{R}^{|\mathcal{S}|}$ is the species-head output of the $k$-th fold.
Rather than calculating the predicted leaf taxon as ${\hat s} = \arg\max_{1\leq i \leq {|\mathcal{S}|}}p_i$, we apply minimum-risk inference following Bayesian decision theory:\cite{duda2001pattern}
\begin{equation}
\hat{s} = \arg\min_{{1\leq i \leq {|\mathcal{S}|}}} \sum_{j=1}^{|\mathcal{S}|} {{p}_j} \cdot D_{ij},
\label{eq:minrisk}
\end{equation}
where $p_j$ is the $j$-th component of the averaged probability vector $\mathbf{p}$, $D_{ij}$ is the pairwise taxonomic distance between leaf taxa $i$ and $j$ (as defined in Equation~\ref{eq:loss}), and the minimization selects the leaf taxon with the lowest expected taxonomic distance under the predicted distribution.
Min-risk biases toward taxonomically central predictions when the model is uncertain, improving mean taxonomic distance by 4.8\% at a small cost to top-1 accuracy ($-1.1\%$)---a favorable trade-off given the evaluation metric.
We additionally apply horizontal-flip test-time augmentation, providing a further 1.2\% improvement.



\section{DATASET}
\label{sec:dataset}

\begin{figure}[t]
\centering
\includegraphics[width=\textwidth]{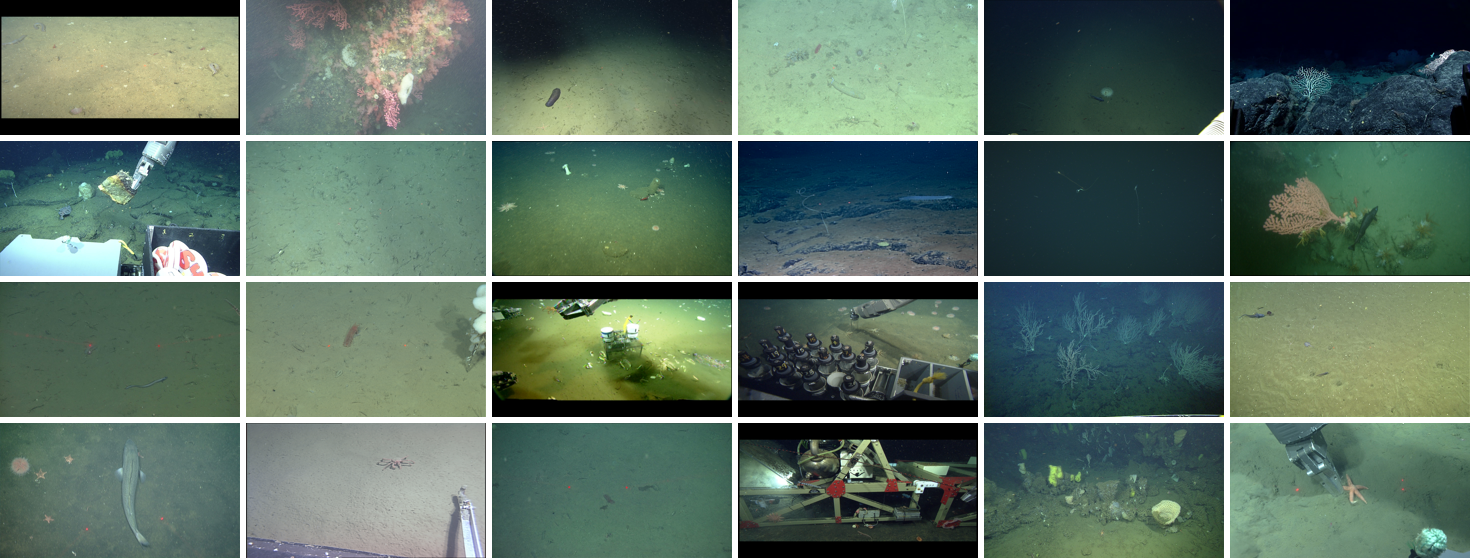}
\caption{Random sample of 24 full-scale training images from the FathomNet 2025 dataset. Images span a wide range of depths, lighting conditions, substrates, and organism types, illustrating the environmental variability that drives distribution shift between training and test sets.}
\label{fig:dataset}
\end{figure}

The FathomNet 2025 dataset\cite{fathomnet2025kaggle} provides 23,699 training ROIs from 79 known categories of marine animals of varying taxonomic ranks (e.g., family, genus, species) plus 1 ``unknown'' category, and 788 test ROIs.
For every category, the taxonomy spans 2~kingdoms (including unknown), 8~phyla, 22~classes, 43~orders, 66~families, 75~genera, and 80~leaf labels.
The species-level classification head outputs a softmax over all $|\mathcal{S}| = 80$ leaves; both the 79 named taxa and ``unknown'' are valid prediction targets, and the leaderboard scores submissions against the same 80-class label space.
Critically, only 33 of the 79 named categories of marine animal are identified down to the species level; the remaining 46 marine animals are identified only to genus, family, order, class, or phylum because finer identification was unavailable.
Coarse-labeled taxa account for 58\% of test organisms.
Taxa are approximately balanced ($\sim$300 instances per category), so the difficulty lies in distribution shift rather than class imbalance.
Figure~\ref{fig:dataset} shows a random sample of full-scale training images illustrating the diversity of underwater scenes and organisms.

\textbf{Distribution shift.}
The competition deliberately sources test images from different geographic locations, depths, and time periods than training images.
Test ROIs are significantly smaller ($84\!\times\!75$~px vs.\ $124\!\times\!115$~px train, scale ratio 0.68$\times$, Cohen's $d = 0.27$), darker (mean ROI brightness 0.379 vs.\ 0.396), and from less densely populated scenes (max~23 annotations/image vs.\ 107 train).
These shifts produce a large gap between validation ($\sim$0.3 mean taxonomic distance) and test performance ($\sim$1.6), though validation remains a reliable relative signal for model selection.
For external generalization assessment, we additionally constructed an independent validation set of 5{,}802 annotations covering 70 of the 79 target taxa, queried from the FathomNet Database API\cite{katija2022fathomnet} and excluding all images appearing in the competition splits.

\section{EXPERIMENTS}
\label{sec:experiments}

\subsection{Implementation Details}

Our primary backbone is DINOv2-Base\cite{oquab2024dinov2} (ViT-B/14, 86M parameters per scale, 351M total including the fusion layer and seven heads).
Fine-tuning uses layer-wise learning rate decay (LLRD)\cite{howard2018ulmfit} with decay factor $\gamma = 0.70$ across 14 parameter groups: the backbone base learning rate is $0.1\times$ the head rate, and each group $i$ is further scaled by $\gamma^{N-1-i}$.
The combined effect is a $\sim$1{,}000$\times$ ratio between the classification head ($3 \times 10^{-4}$) and the earliest parameter group ($\sim$$3 \times 10^{-7}$), preserving robust low-level features while adapting high-level representations.
Optimization uses AdamW\cite{loshchilov2019adamw} with weight decay 0.05, cosine annealing, batch size 12, and 16-bit mixed precision; training is conducted on a single NVIDIA H200 GPU using PyTorch Lightning, with each fold requiring approximately 2~GPU-hours.
At inference, a single fold processes one sample (4 sequential backbone passes) in $\sim$27~ms on an H200 at batch size 1; the 10-fold ensemble scales linearly to $\sim$270~ms/sample with negligible overhead for probability averaging and min-risk decoding ($<$0.1~ms).

\textbf{Design freeze and statistical validation.}
All hyperparameters---backbone choice, fusion method, $\alpha$, head weights, and parent-conditioning---were finalized on cross-validation scores before any test-set evaluation; the chronological order was design, train, freeze, then evaluate.
The ablation tables in this section therefore report test scores retrospectively to characterize generalization, not to guide selection.
The taxonomic distance weight $\alpha = 0.3$ was chosen via a sweep $\alpha \in \{0.0, 0.1, 0.2, 0.3, 0.5\}$ with complete 10-fold ensembles: results are robust over $\alpha \in [0.1, 0.3]$ (within 0.05 of the optimum) and only $\alpha = 0.5$ is significantly worse ($p = 0.02$); pure cross-entropy ($\alpha = 0$) trails the optimum by 2.5\%.
We compute bias-corrected and accelerated (BCa) bootstrap confidence intervals\cite{efron1987bootstrap} with 10,000 resamples on per-sample distances and paired bootstrap tests for pairwise comparisons.
Reported $p$-values are unadjusted; under Holm--Bonferroni correction across the seven ablation rows, the highest-confidence findings (backbone $p = 0.001$, external multi-scale/ensemble/min-risk all $p < 0.001$) remain significant, while marginal results ($p = 0.011$ for min-risk on the 788-sample test) should be interpreted in light of the much tighter external-set evidence.

\subsection{Component-Wise Ablation}
\label{sec:ablation}

Table~\ref{tab:ablation} presents the additive contribution of each component on the 788-sample test set.

\begin{table}[h]
\caption{Component-wise ablation (additive chain). Each row adds one component to the previous configuration.}
\label{tab:ablation}
\centering
\begin{tabular}{clcccc}
\toprule
\# & Configuration & Score & 95\% CI & $\Delta$ & $p$ \\
\midrule
1 & 1$\times$ ROI, CE+LS, single model, argmax & 2.345 & [2.083, 2.622] & --- & --- \\
2 & + Multi-scale (4 scales) & 2.235 & ---$^\dagger$ & $-$4.7\% & --- \\
3 & + Taxonomic distance loss ($\alpha=0.3$) & 2.085 & [1.850, 2.339] & $-$6.7\% & --- \\
4 & + 10-fold ensemble & 1.956 & [1.743, 2.192] & $-$6.2\% & --- \\
5 & + Minimum-risk inference & 1.863 & [1.653, 2.095] & $-$4.8\% & 0.011 \\
6 & + Horizontal-flip TTA & 1.840 & [1.622, 2.074] & $-$1.2\% & 0.450 \\
7 & $-$ Parent conditioning & \best{1.782} & [1.571, 2.013] & $-$3.2\% & 0.214 \\
\bottomrule
\multicolumn{6}{l}{\footnotesize $^\dagger$Single-model checkpoint not preserved; CI unavailable.}
\end{tabular}
\end{table}

The additive chain shows a cumulative 24.0\% improvement over the baseline.
The largest individual gains come from the taxonomy-aware loss ($-$6.7\%) and ensemble averaging ($-$6.2\%); minimum-risk inference is statistically significant ($p = 0.011$).
A scale subset ablation (omitted for brevity) shows that all four scales contribute, with the intermediate 3$\times$ and 5$\times$ context crops more valuable than the full image.
A species-only variant that removes all six auxiliary heads achieves 1.962 versus 1.782 ($+10.1\%$), confirming that auxiliary supervision at coarse levels benefits both the 58\% of coarse-labeled samples and the species-level samples through multi-task regularization.

\textbf{External validation.}
We re-evaluate the same checkpoints on the independent 5{,}802-sample external set described in Section~\ref{sec:dataset} (Table~\ref{tab:external}).
The 7$\times$ larger sample yields BCa CI widths of $\sim$0.12 (vs.\ $\sim$0.49 on the competition test), providing substantially more statistical power for detecting component-wise effects without any test-set probing.

\begin{table}[h]
\caption{External ablation on 5{,}802 independent annotations (same checkpoints as Table~\ref{tab:ablation}). Paired bootstrap $p$ values compare each row to the previous; all add-one comparisons are on the external set.}
\label{tab:external}
\centering
\begin{tabular}{clccc}
\toprule
\# & Configuration & Score & 95\% CI & $p$ vs.\ prev. \\
\midrule
1 & 1$\times$ ROI, CE+LS, single model, argmax & 2.379 & [2.310, 2.451] & --- \\
2 & + Multi-scale (4 scales) & 2.141 & [2.077, 2.206] & $<$0.001 \\
3 & + Taxonomic distance loss ($\alpha=0.3$) & 2.105 & [2.044, 2.170] & 0.106 \\
4 & + 10-fold ensemble & 2.014 & [1.953, 2.076] & $<$0.001 \\
5 & + Minimum-risk inference & \best{1.851} & [1.793, 1.910] & $<$0.001 \\
6 & + Horizontal-flip TTA & 1.859 & [1.802, 1.919] & 0.325 \\
7 & $-$ Parent conditioning & 1.873 & [1.814, 1.932] & 0.331 \\
\bottomrule
\end{tabular}
\end{table}

The three largest components---multi-scale encoding ($-$10.0\%, $p < 0.001$), ensemble averaging ($-$4.3\%, $p < 0.001$), and minimum-risk inference ($-$8.1\%, $p < 0.001$)---all replicate with high significance on the external set.
The taxonomy-aware loss is smaller and not significant here ($p = 0.106$), suggesting its benefit is partially distribution-dependent.
TTA and removing parent conditioning are non-significant on both datasets.

\subsection{Backbone Comparison}
\label{sec:backbone}

Table~\ref{tab:backbone} compares ConvNeXtV2-Base\cite{woo2023convnextv2} (FCMAE pretraining with supervised ImageNet-22K fine-tuning) with DINOv2-Base\cite{oquab2024dinov2} (self-supervised distillation on 142M images).
All other components are identical.

\begin{table}[h]
\caption{Backbone comparison (10-fold ensemble, min-risk + hflip TTA, no parent conditioning).}
\label{tab:backbone}
\centering
\begin{tabular}{lcccc}
\toprule
Backbone & Accuracy$^*$ & Tax.\ Distance & 95\% CI & $p$ \\
\midrule
ConvNeXtV2-Base & 67.8\% & 1.782 & [1.571, 2.013] & --- \\
DINOv2-Base (uniform LR) & --- & 2.225$^\dagger$ & --- & --- \\
\best{DINOv2-Base (LLRD)} & \best{72.2\%} & \best{1.581} & [1.376, 1.805] & 0.001 \\
\bottomrule
\multicolumn{5}{l}{\footnotesize $^*$Top-1 species-level accuracy. $^\dagger$5-fold, argmax inference, uniform backbone LR $3 \times 10^{-5}$.}
\end{tabular}
\end{table}

DINOv2-Base with LLRD achieves 1.581---an 11.3\% improvement over ConvNeXtV2-Base ($p = 0.001$), with DINOv2 outperforming on all 10 individual folds.
This overturns a preliminary finding where DINOv2 with uniform fine-tuning (2.225) appeared substantially worse: applying a uniform learning rate destroys the robust low-level features learned during self-supervised pretraining, a form of catastrophic forgetting.\cite{touvron2022vit}
LLRD preserves these features by fine-tuning lower layers slowly while adapting upper layers more aggressively.
We note that ConvNeXtV2 was fine-tuned with a 2-group schedule rather than per-stage decay, so the 11.3\% gap should be interpreted as an upper bound on the pure architectural advantage; the finding nevertheless underscores that backbone comparisons across CNN/ViT architectures are confounded by fine-tuning recipe.

\subsection{Comparison with Competition Solutions}

\begin{table}[h]
\caption{Comparison with top FathomNet 2025 solutions.}
\label{tab:competition}
\centering
\begin{tabular}{clcl}
\toprule
Rank & Team & Score & Key Technique \\
\midrule
1 & Yonsei+SSL\cite{lee2026matanet} & 1.535 & Multi-context attention + ViT \\
2 & Databaes\cite{kidshock2025} & 1.560 & EfficientNetV2 + uncertainty \\
--- & \best{Ours (post-competition)} & \best{1.581} & DINOv2 LLRD + tax.\ loss + min-risk \\
3 & DalhousieAI\cite{dalhousieai2025} & 2.015 & WideResNet + self-training \\
4 & 911\cite{health9819} & 2.235 & Swin-T + min-risk inference \\
\bottomrule
\end{tabular}
\end{table}

Our team's official competition submission placed 8th on the public leaderboard; the refined framework presented here, incorporating improvements developed after the competition window closed, achieves 1.581---competitive with the 1st-place solution (1.535) without specialized attention mechanisms.
MATANet uses a single DINOv2 ViT-L/14 backbone ($\sim$300M parameters) with cross-attention between multi-scale patch tokens; our approach uses the smaller ViT-B/14 (351M per fold, comparable per-model, with 10-fold ensemble averaging at inference).
The proximity of our concatenation-based result to MATANet's attention-based result suggests that much of the advantage comes from DINOv2's features rather than the attention mechanism itself.

\subsection{Discussion}
\label{sec:discussion}

\textbf{Independent components generalize better.}
A recurring theme is that methods adding learned inter-component dependencies---parent-conditioned heads, attention-based fusion, post-hoc hierarchy repair---tend to overfit or degrade under distribution shift, while independent components allow ensemble averaging to recover from individual errors.
A greedy post-hoc repair strategy that restricts species predictions to descendants of the coarse-level argmax changes only 4.6\% of predictions but worsens taxonomic distance from 1.581 to 1.666 ($+5.4\%$): incorrect coarse predictions force species into the wrong subtree.
Inference-time strategies reinforce this pattern: minimum-risk inference and TTA together provide 6.0\% improvement on the competition test set at zero training cost by exploiting problem structure (the distance matrix) and input symmetry; on the external set, the gain is concentrated in minimum-risk inference ($-$8.1\%, $p < 0.001$), with TTA effectively neutral.

\textbf{Compute-efficient variants.}
DINOv2-Small (ViT-S/14, 22M parameters per backbone, 88M total vs.\ 348M for DINOv2-Base) with independent backbones achieves 1.594---statistically indistinguishable from DINOv2-Base ($p = 0.844$) despite using 74\% fewer backbone parameters.
Sharing a single backbone across all four scales ($\sim$90M total) costs $\sim$8\% performance (1.716) but remains competitive with ConvNeXtV2-Base independent (1.782).
For edge deployment, DINOv2-Small with a shared backbone (24M parameters) achieves 1.713, demonstrating that shift-robust features transfer to the compute-constrained regime.

\textbf{Calibration.}
The 10-fold ensemble is moderately overconfident\cite{guo2017calibration} (mean confidence 85\% vs.\ 72.2\% accuracy).
Cross-validated temperature scaling on out-of-fold predictions (no test leakage) yields $T = 1.88$ and reduces expected calibration error by 42\%; applying this temperature to the locked test predictions improves min-risk performance by 3.8\% (from 1.581 to 1.521---within 1\% of the 1st-place MATANet).
We emphasize that 1.521 is a \emph{local post-hoc evaluation} on the public test set; this configuration was not submitted to the leaderboard during the competition window, and we report 1.581 as the headline result to preserve the chronological design freeze and avoid conflating method performance with post-competition refinement.
External-set calibration was not analyzed in this study---the external set was held out specifically for the unbiased component-wise replication in Table~\ref{tab:external}---and remains a clear next step before deployment.

\textbf{Distance matrix normalization.}
The distance matrix $D$ enters both the loss (as the expectation $\sum_i [\hat{\mathbf{y}}_s]_i \cdot D_{i,y_s}$) and inference (as the risk $\sum_j [\bar{\mathbf{p}}]_j \cdot D_{ij}$).
Normalization by $D_{\max} = 14$ keeps loss magnitudes comparable to cross-entropy and avoids the need to retune $\alpha$ when scaling to deeper hierarchies; replacing it with un-normalized integer distances rescales gradients but leaves the relative cost ordering---and therefore both the optimal $\alpha$ region and the min-risk argmin---unchanged.
Non-uniform per-rank penalties are an alternative we did not explore; this would express domain priorities (e.g., higher cost for cross-class confusions) explicitly in $D$, while keeping the same machinery.

\section{CONCLUSION}

Hierarchical classification of marine species from underwater imagery is complicated by severe domain shift, fine-grained visual similarity, and uneven annotation granularity.
We presented a taxonomy-aware framework that addresses these challenges by aligning both training and inference with the hierarchical evaluation metric, combining a taxonomy-weighted loss, minimum-risk Bayesian inference, multi-scale feature encoding, and independent per-rank classification heads.

Three empirical findings emerge from the component-wise ablation, validated on both the FathomNet 2025 test set and a 7$\times$ larger independent validation set.
First, aligning the loss and inference rule with the evaluation metric produces consistent, significant gains, with minimum-risk inference alone improving performance by 4.8\% at no training cost.
Second, under distribution shift, simple decoupled components (feature concatenation, independent heads) consistently outperform learned dependencies (attention fusion, parent conditioning, post-hoc hierarchy repair).
Third, backbone fine-tuning recipe matters as much as architecture: DINOv2-Base with layer-wise learning rate decay improves over ConvNeXtV2-Base by 11.3\% ($p = 0.001$), but only when fine-tuned to preserve pretrained features.
The full system achieves a mean taxonomic distance of 1.581, within 3\% of the 1st-place solution (1.535) on the FathomNet 2025 public leaderboard.

These lessons---metric-aligned training, robust decoupled design, and careful fine-tuning---are not specific to marine imagery.
Any hierarchical classification task under domain shift where errors carry structured costs, including ATR and ISR applications with vehicle or threat taxonomies, faces the same design trade-offs and can apply the same principles.

\textbf{Limitations and future work.}
The competition test set (788 samples) yields wide confidence intervals ($\sim$0.49); our independent external validation (5{,}802 samples, CIs $\sim$0.12) confirms the core conclusions with substantially more statistical power.
The ConvNeXtV2 comparison did not use stage-wise learning rate decay, so the backbone conclusion may be partially confounded by fine-tuning recipe.
Post-hoc temperature scaling improves the result to 1.521 (within 1\% of first place), suggesting that probability calibration is a natural next refinement for minimum-risk inference.
The species-leaf distance matrix scales as $O(|\mathcal{S}|^2)$ in memory, which is trivial at the current scale ($|\mathcal{S}| = 80$) but would require tree-based dynamic programming for substantially larger hierarchies (e.g., iNaturalist\cite{vanhorn2018inaturalist} with $>$10K species).
Both evaluation sets originate from the FathomNet database; validating on other hierarchical domains and integrating open-set recognition for novel species remain future work.

\acknowledgments

This work was conducted at the Center for Connected Autonomy and Artificial Intelligence (CAAI), Florida Atlantic University.
D.Z.\ is supported by the Department of Defense (DoD) Science, Mathematics, and Research for Transformation (SMART) Scholarship-for-Service Program.
We thank the CAAI graduate students for feedback during manuscript preparation, Shruti Shukla and Pavan Poluri for their initial work on the competition, and the FathomNet team and Kaggle for organizing the competition.
Source code and configuration files will be released at \url{https://github.com/C2A2-at-Florida-Atlantic-University/fathomnet-taxonomy} upon publication.


\end{document}